\documentclass[12pt]{svjour3}
\usepackage[english]{babel}
\usepackage{hyperref}
\usepackage{graphicx, caption}
\usepackage{float}
\usepackage{tabularx}
\usepackage{mathtools}
\usepackage{multirow}
\usepackage{amssymb}
\usepackage{lscape}
\usepackage{xfrac}
\usepackage{eurosym}
\usepackage{stmaryrd}
\usepackage{subcaption}

\usepackage{tikz}
\usetikzlibrary{calc}
\usetikzlibrary{decorations.pathreplacing,calligraphy,calc}
\tikzstyle{node}=[draw,circle]
\tikzstyle{arc}=[>=latex,->]
\usepackage{graphicx}
\usepackage{comment}
\usepackage{color}
\usepackage{geometry}
\makeatletter

\pgfdeclaredecoration{my calligraphic brace}{brace}
{
  \state{brace}[width=+\pgfdecoratedremainingdistance,next state=final]
  {
    \pgfsyssoftpath@setcurrentpath{\pgfutil@empty}
    \pgfpathmoveto{\pgfpointorigin}
    \pgfpathmoveto{\pgfqpoint{1mm}{-2mm}} 
    \pgfpathcurveto
    {\pgfqpoint{.15\pgfdecorationsegmentamplitude}{.3\pgfdecorationsegmentamplitude}}
    {\pgfqpoint{.5\pgfdecorationsegmentamplitude}{.5\pgfdecorationsegmentamplitude}}
    {\pgfqpoint{\pgfdecorationsegmentamplitude}{.5\pgfdecorationsegmentamplitude}}
    {
      \pgftransformxshift{+\pgfdecorationsegmentaspect\pgfdecoratedremainingdistance}
      \pgfpathlineto{\pgfqpoint{-\pgfdecorationsegmentamplitude}{.5\pgfdecorationsegmentamplitude}}
      \pgfpathcurveto
      {\pgfqpoint{-.5\pgfdecorationsegmentamplitude}{.5\pgfdecorationsegmentamplitude}}
      {\pgfqpoint{-.15\pgfdecorationsegmentamplitude}{.7\pgfdecorationsegmentamplitude}}
      {\pgfqpoint{0\pgfdecorationsegmentamplitude}{1\pgfdecorationsegmentamplitude}}
      \pgfpathmoveto{\pgfqpoint{0\pgfdecorationsegmentamplitude}{1\pgfdecorationsegmentamplitude}}
      \pgfpathcurveto
      {\pgfqpoint{.15\pgfdecorationsegmentamplitude}{.7\pgfdecorationsegmentamplitude}}
      {\pgfqpoint{.5\pgfdecorationsegmentamplitude}{.5\pgfdecorationsegmentamplitude}}
      {\pgfqpoint{\pgfdecorationsegmentamplitude}{.5\pgfdecorationsegmentamplitude}}
    }
    {
      \pgftransformxshift{+\pgfdecoratedremainingdistance}
      \pgfpathlineto{\pgfqpoint{-\pgfdecorationsegmentamplitude}{.5\pgfdecorationsegmentamplitude}}
      \pgfpathcurveto
      {\pgfqpoint{-.5\pgfdecorationsegmentamplitude}{.5\pgfdecorationsegmentamplitude}}
      {\pgfqpoint{-.15\pgfdecorationsegmentamplitude}{.3\pgfdecorationsegmentamplitude}}
      {\pgfqpoint{-1mm}{-2mm}}  
    }
    \tikzset{
      taper width=.5\pgflinewidth,
      taper
    }%
    \pgfsyssoftpath@getcurrentpath\cal@tmp@path
    \MakeSPathList{calligraphy path}{\cal@tmp@path}%
    \SPathListPrepare{calligraphy path}%
    \CalligraphyPathCreate{calligraphy path}{copperplate}%
  }
  \state{final}{}
}
\makeatother

\begin{document}

\title{Learning MR-Sort Models from Non-Monotone Data}

% \author{P. Minoungou, V. Mousseau, W. Ouerdane, and P. Scotton}
% \institute{
% Pegdwendé Minoungou \at MICS, CentraleSup\'elec, Universit\'e Paris-Saclay, Gif-sur-Yvette, France \\ IBM, Saclay, France \\ \email{pegdwende.minoungou@centralesupelec.fr,Pegdwende.Stephane.Minoungou@ibm.com}  \and
% Vincent Mousseau \at MICS, CentraleSup\'elec, Universit\'e Paris-Saclay, Gif-sur-Yvette, France \\ \email{vincent.mousseau@centralesupelec.fr}  \and
% Wassila Ouerdane \at MICS, CentraleSup\'elec, Universit\'e Paris-Saclay, Gif-sur-Yvette, France \\ \email{wassila.ouerdane@centralesupelec.fr} \and
% Paolo Scotton \at IBM Research Europe - Zurich, Switzerland \\ \email{psc@zurich.ibm.com} }

% \authorrunning{P. Minoungou, V. Mousseau, W. Ouerdane, and P. Scotton}

\author{Pegdwendé Minoungou (1,2) \and
Vincent Mousseau (1) \and
Wassila Ouerdane (1) \and
Paolo Scotton (3)}
\authorrunning{P. Minoungou, V. Mousseau, W. Ouerdane, and P. Scotton}
\institute{1 MICS, CentraleSup\'elec, Universit\'e Paris-Saclay, Gif-sur-Yvette, France
\email{\{pegdwende.minoungou, vincent.mousseau, wassila.ouerdane\}@centralesupelec.fr}\\
2 IBM, Saclay, France 
\email{Pegdwende.Stephane.Minoungou@ibm.com}\\
3 IBM Research Europe - Zurich, Switzerland 
\email{psc@zurich.ibm.com}}

\maketitle

\begin{abstract}
The Majority Rule Sorting (MR-Sort) method assigns alternatives evaluated on multiple criteria to one of the predefined ordered categories. The Inverse MR-Sort problem (Inv-MR-Sort) computes MR-Sort parameters that match a dataset. Existing learning algorithms for Inv-MR-Sort consider monotone preferences on criteria. We extend this problem to the case where the preferences on criteria are not necessarily monotone, but possibly single-peaked (or single-valley). We propose a mixed-integer programming based algorithm that learns the preferences on criteria together with the other MR-Sort parameters from the training data. We investigate the performance of the algorithm using numerical experiments and we illustrate its use on a real-world case study.
\end{abstract}

%============================
\section{Introduction}
\label{sec:intro}
%============================

In this paper we consider multiple criteria sorting problems in which alternatives evaluated on several criteria are to be assigned to one of the pre-defined ordered categories $C^1$, $C^2$, ..., $C^p$, $C^1$ being the worst and $C^p$ being the best category. \\[-3mm]

Many multiple criteria methods have been proposed in the literature~\cite{DoumposZopounidis2002}~\cite{zopoudoupos-review}. We are interested in a pairwise comparison based method: the Non-Compensatory Sorting model~ \cite{bouyssou2007a}~\cite{bouyssou2007b} (NCS). NCS assigns alternatives to categories based on the way alternatives compare to external profiles representing frontiers between consecutive categories, and can be viewed as an axiomatic formulation of the Electre Tri method~\cite{Roy1991}. More specifically, we consider a particular case of NCS in which the importance of criteria is additively represented using weights: the Majority Rule Sorting~\cite{leroy2011learning} (MR-Sort). \\[-3mm]

In real-world decision problems involving multiple criteria sorting, the implementation of a sorting model requires eliciting the decision-maker's (DM) preferences and adequately representing her preferences by setting appropriate values for the preference-related parameters. It is usual to elicit the sorting model parameters indirectly from a set of assignment examples, i.e., a set of alternatives with the associated desired category. Such a preference learning approach has been developed for MR-Sort (Inv-MR-Sort~\cite{leroy2011learning}~\cite{SobrieMousseauPirlot2019}), and makes it possible to compute MR-Sort parameters that best fit a learning set provided by the DM.\\[-3mm]

The approach considers criteria involving monotone preferences (criteria to be maximized or minimized). This applies in the context of Multiple Criteria Decision Aid (MCDA) in which the decision problem is structured and carefully crafted by the DM. In contrast, we are interested in this paper in applications in which the evaluation of alternatives on criteria does not necessarily induce monotone preferences. We illustrate hereafter such a situation with two examples. \\[-3mm]

Example 1: Consider a veterinary problem in cattle production in which a new cattle disease should be diagnosed based on symptoms: each cattle should be classified as having or not having the disease. New scientific evidence has indicated that, in addition to usual symptoms, the presence of substance A in the blood of the animal can be predictive, but there is no indication on how the level of substance A should be considered. Does a high- or low-level, or a level between bounds of substance A indicate a sick cattle ? 
The veterinarians' union has gathered a large number of cases, and wants to use this data to define a sorting model based on usual symptom criteria and the level of substance A in the blood of the animal. Hence, the sorting model should be inferred from the data, even if the way to account for the Substance A level is unknown. \\[-3mm]

Example 2: A computer-products retail company is distributing a new Windows\footnote{Windows is a trademark of Microsoft Corporation in the United States, other countries, or both.} tablet, and wants to send targeted marketing emails to clients who might be interested in this new product. To do so, clients are to be sorted into two categories: \textit{potential buyer} and  \textit{not interested}. 
To avoid spamming, only clients in the former category will receive a telephone call. To sort clients, four client characteristics are considered as criteria, all of them being homogeneous to a currency e.g. \euro{} : the spend over the last year of \textit{(i)} Windows PC, \textit{(ii) }Pack Office, \textit{(iii)} Linux\footnote{The registered trademark Linux® is used pursuant to a sublicense from the Linux Foundation, the exclusive licensee of Linus Torvalds, owner of the mark on a world wide basis.} PC, and \textit{(iv)} Dual boot PC. 
As the company is advertising a new Windows tablet, both of the first two criteria are to be maximized (the more a client buys Windows PCs and Pack Office, the more he/she is interested in products with a Windows system), and the third criterion is to be minimized (the more a client buys Linux PCs, the less he/she is interested in products with a Windows system). The marketing manager is convinced that the last criterion should be taken into account, but does not know if it should be maximized, minimized, if preferences are single peaked; a subset of clients has been partitioned into not interested/potential buyer. Based on this dataset, the goal is to simultaneously learn the classifier parameters and the preference direction for the last criterion. \\[-3mm]

In the previous examples, it is not clear for the DM how to account for some of the data (level of substance A in blood, Dual boot PC turnover) on the classification of alternatives (cattle, client). In this paper, we assume that evaluations on criteria should be either maximized, minimized or correspond to single-peaked (or single-valley) preferences. We propose a mixed-integer mathematical programming (MIP) approach to learn the MR-Sort parameters and criteria type (gain, cost, single-peaked, or single-valley) from assignment examples.\\ [-3mm]

The paper is organised as follows. Section~\ref{sec:related_work} reviews the existing works in the field of MCDA that consider criteria that are not necessarily monotone. The NCS and MR-Sort methods are presented and extended to the case of single-peaked (single-valley) criteria in Section~\ref{sec:mrsort_mrsortsp}. In Section~\ref{sec:inv-mrsort} we specify the Inv-MR-Sort problem in the presence of single-peaked criteria, and a MIP based algorithm is proposed in Section~\ref{sec:exact_resolution}. Section~\ref{sec:experiments_results} presents the performance of the algorithm on a generated dataset and on a real-world case study. The last section groups conclusions and further research issues.

%============================
\section{Related work}
\label{sec:related_work}
%============================

In Multiple Criteria Decision Aid (MCDA), preference learning methods require a preference order on criteria. Such preference order on criteria directly results from the fact that alternative evaluations/scores correspond to performances that are to be maximized (profit criterion) or minimized (cost criterion), which leads to monotone preference data. In multicriteria sorting problems, this boils down to a higher evaluation on a profit criterion (on a cost criterion, respectively) favouring an assignment to a higher category (to a lower category, respectively). 

However, there are numerous situations in which the evaluation on criteria is not related to category assignment in a monotone way. Such situations are indeed considered in the induction of monotone classification rules from data. 

Classification methods in the field of machine learning usually account for attributes (features) which are not supposed to be monotone. Some specialized methods have been proposed to consider monotone feature~\cite{GutierrezG16}~\cite{Cano2019168} for decision trees \cite{Feelders10} or for decision rules~\cite{GrecoMS01}. Some of these approaches have been extended to partially monotone data~\cite{Pei2018104}~\cite{Wang2015172}. Blaszczy\`{n}ski et al. in~\cite{BLASZCZYNSKI2012284} present a non invasive transformation applied to a dominance-based rough set approach to discover monotonicity relationships (positively/negatively global/local monotonicities) between attributes and the decision considering non-ordinal and ordinal classification problems. With their proposed transformation applied to non-monotone data, they are able to deduce laws with interval conditions on attributes because these attributes are positively monotone in one part of the evaluation space and negatively monotone in the other part. \\[-3mm]

In the context of the multicriteria decision aid, several preference learning/disaggregation approaches consider non-monotone preferences on criteria. To the best of our knowledge, however, almost all these contributions consider a utility-based preference model, in which non-monotone attributes are represented using non-monotone marginal utility functions.  \\[-3mm]

Historically, Despotis and Zopounidis~\cite{Despotis1995} are the first to consider single peaked value functions with an additive piece-wise linear  model. The UTA-NM method proposed in~\cite{Kliegr2009UTANME} allows for non-monotone marginals and prevents over-fitting by introducing a shape penalization. 

Also in the context of an additive utility model, Eckhardt and Klieger~\cite{Eckhardt2012PreprocessingAF} define a heuristic pre-processing technique to transform arbitrary attributes input into a space monotone w.r.t. the Decision Maker's (DM) preferences. Another contribution proposed by Doumpos~\cite{Doumpos2012} proposes a heuristic approach to learn a non-monotone additive value-based sorting model from data. 

Liu \textit{et al.}~\cite{LIU20191071} model sorting with a piece-wise linear additive sorting model, using a regularization framework to limit non-monotonicity. Guo\textit{ et al. }\cite{GUO2019} propose a progressive preference elicitation for multicriteria sorting using a utility model with non-monotone attributes. A framework to rank alternatives with a utility model using slope variation restrictions for marginals is proposed in~\cite{GHADERI2017}. 

Based on a mixed-integer program, \cite{KADZINSKI202060,Kadzinski2021a} proposes to disaggregate an additive piece-wise linear sorting model with different types of monotone (increasing, decreasing) and non-monotone (single peaked, single caved) marginal value functions.

Recently some contributions aim at inferring non-compensatory sorting models involving non-monotone criteria from data. Sobrie et al.~\cite{sobrie2016ASA} consider single-peaked preferences when learning an MR-Sort model in a medical application. In contrast, \cite{minoungouDA2PL2020} proposed a heuristic to learn an MR-Sort model and criteria preference directions from data.

%%%%%%%%%%%%%%%%%%%%%%%%%%%%%%%%%%%%%%%%%%%%%%%%%
\section{NCS, MR-Sort, and single-peaked preferences}
\label{sec:mrsort_mrsortsp}
%%%%%%%%%%%%%%%%%%%%%%%%%%%%%%%%%%%%%%%%%%%%%%%%%
\subsection{NCS: Non-compensatory Sorting}
%%%%%%%%%%%%%%%%%%%%%%%%%%%%%%%%%%%%%%%%%%%%%%%%%

Non-compensatory Sorting (NCS)~\cite{bouyssou2007a,bouyssou2007b} is an MCDA sorting model originating from the ELECTRE TRI method~\cite{figueira2005electre}. NCS can be intuitively formulated as follows: an alternative is assigned to a category if \emph{(i)} it is better than the lower limit of the category on a sufficiently strong subset of criteria, and \emph{(ii)} this is not the case when comparing the alternative to the upper limit of the category. \\[-3mm]

Consider the simplest case involving 2 categories Good ($\mathcal{G}$) and  Bad ($\mathcal{B})$ with the following notations. We denote $X_i$ the set of possible values on criterion $i$, $i \in \mathcal{N} =  \{1, \ldots, n\}$. Hence,  $X = \prod_{i \in \mathcal{N}}  X_i$ represents the set of alternatives to be sorted. We denote $\mathcal{A}_i \subseteq X_i$ the set of approved values on criterion $i \in \mathcal{N}$. Approved values on criterion $i$ ($x_i \in \mathcal{A}_i$) correspond to values contributing to the assignment of an alternative to category $\mathcal{G}$. In order to assign alternative $a$ to category $\mathcal{G}$, $a$ should have approved values on a subset of criteria which is ``sufficiently strong''. The set $\mathcal{F}\subseteq 2^{\mathcal{N}}$ contains the ``sufficiently strong'' subsets of criteria; it is a subset of $2^{\mathcal{N}}$ up-closed by inclusion. In this perspective, the NCS assignment rule can be expressed as follows:
%\vspace*{-1mm}
\begin{equation}
    x \in \mathcal{G} \quad \text{iff} \quad \{i \in \mathcal{N} : x_i \in \mathcal{A}_i \} \in \mathcal{F}, \;\;\; \forall x  \in X
    \label{NCS-rule-2-cat}
\end{equation}

With more than two categories, we consider an ordered set of $p$ categories $C^p \rhd  \dots \rhd C^h \rhd \dots \rhd C^1$, where $\rhd$ denotes the order on categories.  Sets of approved values $\mathcal{A}_i^h \subseteq X_i$ on criterion $i$ ($i \in \mathcal{N}$) are defined with respect to a category $h$ ($h=1..p-1$), and should be defined as embedded sets such that $\mathcal{A}^1_i \supseteq \mathcal{A}^2_i \supseteq ... \supseteq \mathcal{A}^{p-1}_i$. Analogously, sets of sufficiently strong criteria coalitions are relative to a category $h$, and are embedded as follows: $\mathcal{F}^1 \subseteq \mathcal{F}^2 \subseteq ... \subseteq \mathcal{F}^{p-1} $. The assignment rule is defined, for all $x \in X$, as:

\begin{equation}
    x \in C^h \quad \text{iff} \quad \{i \in \mathcal{N} : x_i \in \mathcal{A}^h_i \} \in \mathcal{F}^h \mbox{ and } \{i \in \mathcal{N} : x_i  \in \mathcal{A}^{h+1}_i\} \notin \mathcal{F}^{h+1}
    \label{NCS-rule-n-cat}
\end{equation}

\medskip
A particular case of NCS corresponds to the MR-Sort rule~\cite{leroy2011learning}. When the families of sufficient coalitions are all equal $\mathcal{F}^1 = ... = \mathcal{F}^{p-1} = \mathcal{F}$ and is defined using additive weights attached to criteria, and a threshold: $\mathcal{F} = \{ F \subseteq \mathcal{N} : \sum_{i \in F} w_i \ge \lambda \}$,  with $w_i\ge 0$, $\sum_{i} w_i=1$, and $\lambda \in  [0,1]$. Moreover, we consider that $X_i\subset \mathbb{R}$ and the order on $\mathbb{R}$ induces a complete pre-order $\succcurlyeq_i$ on $X_i$. Hence, the sets of approved values on criterion $i$, $\mathcal{A}_i^h \subseteq X_i$ ($i \in \mathcal{N}, h=1 ... p-1$) are defined by $\succcurlyeq_i$ and $b_i^h\in X_i$ the minimal approved value in $X_i$ at level $h$: $\mathcal{A}_i^h = \{x_i \in X_i : x_i \succcurlyeq_i b_i^h \}$. In this way, $b^h=(b_1^h, \ldots, b_n^h)$ is interpreted as the frontier between categories $C^{h-1}$ and $C^{h}$. Therefore, the MR-Sort rule can be expressed as:

\begin{equation}
    x \in C^h \quad \text{iff} \quad \sum_{i:x_i \ge b_i^h} w_i \ge \lambda \mbox{ and } \sum_{i:x_i \ge b_i^{h+1}} w_i < \lambda
    \label{MR-Sort-rule}
\end{equation}

It should be emphasized that in the above definition of the MR-Sort rule, the approved sets $\mathcal{A}_i^h$ can be defined using $b^h \in X$, which are interpreted as frontiers between consecutive categories, only if preferences $\succcurlyeq_i$ on criterion $i$ are supposed to be monotone, and a criterion can be either defined as a \emph{gain} or a \emph{cost} criterion:
\begin{definition}
        a criterion $i \in {\mathcal N}$ is:
        \begin{itemize}
            \item a gain criterion: when $x_i \ge_i x_i^\prime \; \Rightarrow x_i \succcurlyeq_i x_i^\prime$
            \item a cost criterion: when $x_i \le_i x_i^\prime \; \Rightarrow x_i \succcurlyeq_i x_i^\prime$
        \end{itemize}
        \label{gain-cost-criteria}
\end{definition}

Indeed, in the case of a gain criterion, we have $x_i \in \mathcal{A}_i^h$ and $x_i^\prime >_i x_i \Rightarrow x_i^\prime \in \mathcal{A}_i^h$, and $x_i \notin \mathcal{A}_i^h$ and $x_i >_i x_i^\prime \Rightarrow x_i^\prime \notin \mathcal{A}_i^h$. Therefore, $\mathcal{A}_i^h$ is specified by $b_i^h\in X_i$: $\mathcal{A}_i^h=\{ x_i \in X_i : x_i \ge b_i^h \}$. In the case of a cost criterion, we have $x_i \in \mathcal{A}_i^h$ and $x_i^\prime <_i x_i \Rightarrow x_i^\prime \in \mathcal{A}_i^h$, and $x_i \notin \mathcal{A}_i^h$ and $x_i <_i x_i^\prime \Rightarrow x_i^\prime \notin \mathcal{A}_i^h$. Therefore, $\mathcal{A}_i^h$ is specified by $b_i\in X_i$: $\mathcal{A}_i^h=\{ x_i \in X_i : x_i \le b_i^h \}$. We study hereafter the MR-Sort rule in the case of single-peaked preferences~\cite{black1948} .

%%%%%%%%%%%%%%%%%%%%%%%%%%%%%%%%%%%%%%%%%%%%%%%%%
\subsection{Single-peaked and single-valley preferences}
\label{subsec:inv-mrsort-single-peaked}
%%%%%%%%%%%%%%%%%%%%%%%%%%%%%%%%%%%%%%%%%%%%%%%%%
In the following, we suppose $X_i=[min_i,max_i] \subset \mathbb{R}$, and denotes $>_i$ as the order on $X_i$ induced from the order on $\mathbb{R}$. 
\begin{definition}
  Preferences  $\succcurlyeq_i$ on criterion $i$ are:
  \vspace*{-2mm}
  \begin{itemize}
      \item single-peaked preferences with respect to $>_i$ iff there exist $p_i \in X_i$ such that: $x_i \le_i y_i <_i p_i \Rightarrow p_i \succ_i y_i \succcurlyeq_i x_i$, and  $p_i <_i x_i \le_i y_i \Rightarrow p_i \succ_i x_i \succcurlyeq_i y_i$
      \item single-valley preferences with respect to $>_i$ iff there exist $p_i \in X_i$ such that: $x_i \le_i y_i <_i p_i \Rightarrow p_i \succ_i x_i \succcurlyeq_i y_i$, and  $p_i <_i x_i \le_i y_i \Rightarrow p_i \succ_i y_i \succcurlyeq_i x_i$
  \end{itemize}
  \label{single-peaked-preferences}
\end{definition} 

From an MCDA perspective, single-peaked preferences (single-valley, respectively) can be interpreted as a gain criterion to be maximized (a cost criterion to be minimized, respectively) bellow the peak $p_i$, and as a cost criterion to be minimized (a gain criterion to be maximized, respectively) above the peak $p_i$. Note also that single-peaked and single-valley preferences embrace the case of gain and cost criteria: a gain criterion corresponds to single-peaked preferences when $p_i=max_i$ or single-valley preferences with $p_i=min_i$, and a cost criterion corresponds to single-peaked preferences when $p_i=min_i$ or single-valley preferences with $p_i=max_i$.\\[-3mm]

When considering MR-Sort with single-peaked criteria, approved sets can not be represented using frontiers between consecutive categories. However, approved sets should be compatible with preferences, i.e. such that:
\begin{equation}
    \left \{
    \begin{array}{l}
          x_i \in \mathcal{A}_i^h \mbox{ and } x_i^\prime \succcurlyeq_i x_i \Rightarrow x_i^\prime \in \mathcal{A}_i^h \\
         x_i \notin \mathcal{A}_i^h \mbox{and } x_i \succcurlyeq_i x_i^\prime \Rightarrow x_i^\prime \notin \mathcal{A}_i^h 
    \end{array}
    \right .
    \label{approved-sets}
\end{equation}
\noindent In case of a single-peaked criterion with peak $p_i$, we have:
\begin{equation}
    \left \{
    \begin{array}{l}
          x_i \in \mathcal{A}_i^h \mbox{ and } p_i <_i x_i^\prime <_i x_i \Rightarrow x_i^\prime \in \mathcal{A}_i^h\\
          x_i \in \mathcal{A}_i^h \mbox{ and } x_i <_i x_i^\prime  <_i p_i  \Rightarrow x_i^\prime \in \mathcal{A}_i^h\\
          x_i \notin \mathcal{A}_i^h \mbox{ and } p_i <_i x_i <_i x_i^\prime \Rightarrow x_i^\prime \notin \mathcal{A}_i^h\\
          x_i \notin \mathcal{A}_i^h \mbox{ and } x_i^\prime <_i x_i  <_i p_i \Rightarrow x_i^\prime \notin \mathcal{A}_i^h\\
    \end{array}
    \right .
    \label{single-peaked-approved-sets}
\end{equation}
Therefore, it appears that with a single-peaked criterion with peak $p_i$, the approved sets $\mathcal{A}_i^h$ can be specified by two thresholds $\overline{b}_i^h,\underline{b}_i^h\in X_i$ with $\underline{b}_i^h < p_i < \overline{b}_i^h$ defining an interval of approved values: $\mathcal{A}_i^h=  [ \underline{b}_i^h, \overline{b}_i^h ]$. Analogously,  for a single-valley criterion with peak $p_i$, the approved sets $\mathcal{A}_i^h$ can be specified using  $\overline{b}_i^h,\underline{b}_i^h\in X_i$ (such that $\underline{b}_i^h < p_i < \overline{b}_i^h$) as  $\mathcal{A}_i^h= X_i \, \setminus \, ] \underline{b}_i^h, \overline{b}_i^h [$.   \\[-3mm]

Given a single-peaked criterion $i$, for which an approved set is defined by the interval $\mathcal{A}_i^h= [\underline{b}_i^h, \overline{b}_i^h ]$, let us consider the function $\phi_i: X_i \rightarrow X_i$ defined by $\phi_i(x_i)= | x_i-\frac{\underline{b}_i^h + \overline{b}_i^h}{2} |$; then, the approved set can be conveniently rewritten as : $\mathcal{A}_i^h=  \{ x_i \in X_i: \phi(x_i) \le \frac{\underline{b}_i^h - \overline{b}_i^h}{2}  \}$. In other words, when defining approved sets, a single-peaked criterion can be re-encoded into a cost criterion, evaluating alternatives as the distance to the middle of the interval $[\underline{b}_i^h, \overline{b}_i^h ]$, and a frontier corresponding to half the width of this interval. \\[-3mm]

Analogously, given a single-valley criterion $i$ for which approved sets are defined by the interval $\mathcal{A}_i^h= X_i \, \setminus \, ]\underline{b}_i^h, \overline{b}_i^h  [$, using the same function $\phi_i$, the approved set can be conveniently rewritten as : $\mathcal{A}_i^h= \{ x_i : \phi(x_i) \ge \frac{\underline{b}_i^h - \overline{b}_i^h}{2} \}$. Hence, when defining approved sets, a single-valley criterion can be re-encoded into a gain criterion, evaluating alternatives as the distance to the middle of the interval $ [\underline{b}_i^h, \overline{b}_i^h  ]$, and a frontier corresponding to half the width of this interval.

%===============================
\section{Inv-MR-Sort: Learning an MR-Sort model from assignment examples}
\label{sec:inv-mrsort}
%===============================

MR-Sort preference parameters, e.g. weights, majority level, and limit profiles, can be either initialized by the ``end-user'', i.e. the decision maker, or learned through a set of assignment examples called a learning set. We are focusing on the learning approach. The aim is to find the MR-Sort parameters that ``best'' fit the learning set.

 We consider as input a learning set, denoted $L$, composed of assignment examples. Here, an  \emph{assignment example} refers to an alternative $a \in \ A^\star \subset X$, and a desired category $c(a) \in \{ 1, \ldots, p \}$. In our context, the determination of the MR-sort parameters' value relies on the resolution of a mathematical learning program based on assignment examples: the $\textit{Inv-MR-Sort}$ problem takes as input a learning set $L$ and computes  weights ($w_i, i \in {\mathcal N}$), majority level ($\lambda$), and limit profiles ($b_h, h=1..p-1$) that best restore $L$, i.e. that maximizes the number of correct assignments.\\[-3mm]

This learning approach -- also referred to as preference disaggregation -- has been previously considered in the literature. In particular, \cite{mousseau1998},\cite{zheng2014COR} learned the ELECTRE TRI parameters using a nonlinear programming formulation, while \cite{DOUMPOS2009496} propose an evolutionary approach to do so.
Later, a more amenable model, the MR-Sort -- which derives from the ELECTRE TRI method and requires less parameters than ELECTRE TRI -- was introduced by Leroy et al. in~\cite{leroy2011learning}. They proposed a MIP implementation for solving the $\textit{Inv-MR-Sort}$ problem, while Sobrie et al.~\cite{sobrie2016} tackled it with a metaheuristic and Belahcene et al.~\cite{article:Belahcene2018COR} with a Boolean satisfiability (SAT) formulation.
Other authors proposed approaches to infer MR-Sort incorporating veto phenomenon~\cite{MeyerOlteanu_2017}, and imprecise/missing evaluations~\cite{MeyerOlteanu_2019}, and~\cite{Nefla_2019} presented an interactive elicitation for the learning of MR-Sort parameters with given profiles values. Recently~\cite{KADZINSKI2020} proposed an enriched preference modelling framework which accounts for different types of input.
Lastly,~\cite{minoungouDA2PL2020} proposed an extension of Sobrie's algorithm for solving the $\textit{Inv-MR-Sort}$ problem with latent preference directions, i.e.  considering criteria whose preference direction, in terms of gain/cost, is not known beforehand.  \\[-3mm]

In this paper, we aim at extending the resolution of the Inv-MR-Sort problem to the case where each of the criteria can be either a cost criterion, a gain criterion, a single-peaked criterion, or a single-valley criterion.

%===============================
\section{Exact resolution of Inv-MR-Sort with single-peaked criteria}
\label{sec:exact_resolution}
%===============================

In this section, we present a Mixed Integer Programming (MIP) formulation to solve the $\textit{Inv-MR-Sort}$ problem when each criterion  can either be  a cost, gain, single-peaked, or single-valley criterion. More precisely, the resolution will take as input a learning set containing assignment examples, and computes:
\vspace*{-1.5mm}
\begin{itemize}
    \item the nature of each criterion (either cost, gain, single-peaked, or single-valley criterion),
    \item the weights attached to criteria $w_i$, and an associated majority level $\lambda$,
    \item the frontier between category $C^h$ and $C^{h+1}$, i.e. -- as defined in Section~\ref{sec:mrsort_mrsortsp} -- the value $b_i^h$ such that if criterion $i$ is a cost or a gain criterion, and the interval $[\underline{b}_i^h, \overline{b}_i^h]$  if criterion $i$ is a single-peaked or single-valley criterion.
\end{itemize} 

For the sake or simplicity, we describe the mathematical formulation in the case of two categories; the extension to more than two categories is discussed in Section~\ref{subsec:more-than-2-categories}.\\[-3mm]

Let us consider a learning set $L$, provided by the Decision Maker, containing assignment examples corresponding to a set of reference alternatives $A^{*}=A^{*1}\cup A^{*2}$ partitioned into 2 subsets $A^{*1}=\{a^j \in A^{*}:a^j \in C^1\}$ and $A^{*2}=\{a^j \in A^{*} :a^j \in C^2\}$. We denote by  $J^{*}$, $J^{*1}$, and $J^{*2}$ the  indices $j$ of alternatives contained in $A^{*}$, $A^{*1}$, and $A^{*2}$, respectively.\\[-3mm] 

In the MIP formulation proposed in this section, we represent single-peaked or single-valley criteria only. As discussed in Section~\ref{subsec:inv-mrsort-single-peaked}, this is not restrictive because cost and gain criteria are particular cases of single-peaked (or single-valley) criteria, with a peak corresponding to the end points of the evaluation scale.\\[-3mm]

%----------------------------
\subsection{Variables and constraints related to approved sets and profiles}
\label{subsec:dec_var_profiles}
%----------------------------

Suppose that criterion $i$ is single-peaked and that the set of approved values is defined by $\mathcal{A}_i=[\underline{b}_i, \overline{b}_i]$. Let us denote $b^{\bot}_i = \frac{\overline{b}_i + \underline{b}_i}{2}$ as the middle of the interval of approved values. Consider an alternative $a^j\in A^*$ in the learning set; its evaluation on criterion $i$ is approved (i.e, $a_i^j \in \mathcal{A}_i$) if $a_i^j \in [\underline{b}_i, \overline{b}_i]$. The condition $|a_i^j-b^{\bot}_i| \le \frac{\overline{b}_i - \underline{b}_i}{2}$ guaranties that $a_i^j \in [\underline{b}_i, \overline{b}_i]$. This allows the set $\mathcal{A}_i$ to be rewritten as $\mathcal{A}_i = \{ x_i\in X_i : |x_i-b^{\bot}_i| \le \frac{\overline{b}_i - \underline{b}_i}{2} \}$. 

To test whether $a_i^j \in \mathcal{A}_i$, we define $\alpha^j_i= a^j_i - b^{\bot}_i$ such that  $a_i^j \in \mathcal{A}_i \Leftrightarrow |\alpha^j_i| \le \frac{\overline{b}_i - \underline{b}_i}{2}$. 
In other words, we re-encode criterion $i$ as a cost criterion representing the distance to $b^{\bot}_i$; $b_i=\frac{\overline{b}_i - \underline{b}_i}{2}$ being the frontier of this criterion, i.e. half the interval $[\underline{b}_i, \overline{b}_i]$. Hence, in our formulation, the sets $\mathcal{A}_i$ is defined using two variables: $b^{\bot}_i$ representing the middle of the interval $[\underline{b}_i, \overline{b}_i]$, and $b_i$ allowing us to define $\mathcal{A}_i$ as $\mathcal{A}_i = \{ x_i\in X_i : |x_i-b^{\bot}_i| \le b_i \}$\\[-3mm]

In order to linearize the expression $|\alpha^j_i|=|a^j_i - b^{\bot}_i|$ in the MIP formulation, we consider two positive variables $\alpha^{j+}_i$ , $\alpha^{j-}_i$ and binary variables $\beta^j_i$ verifying constraints (\ref{eq:abs-beta2})-(\ref{eq:abs-beta4}), where $M$ is an arbitrary large positive value. Constraints  (\ref{eq:abs-beta3}) and (\ref{eq:abs-beta4}) ensure that at least one variable among $\alpha^{j+}_i$ and $\alpha^{j-}_i$ is null.
\begin{subequations}
\begin{align}
& \alpha^j_i = a^j_i - b^{\bot}_i = \alpha^{j+}_i - \alpha^{j-}_i \label{eq:abs-beta2} \\
& 0 \leq \alpha^{j+}_i \leq  \beta^j_i M \label{eq:abs-beta3} \\
& 0 \leq \alpha^{j-}_i \leq (1-\beta^j_i) M \label{eq:abs-beta4} 
\end{align}
\end{subequations}

\newpage

Let $\delta_{ij}\in \{0,1\}, \; i \in {\mathcal N}, \; j \in J^{*}$, be binary variables expressing the membership of evaluation $a^j_i$ in the approved set~$\mathcal{A}_i$ ($\delta_{ij}=1 \Leftrightarrow a^j_i \in \mathcal{A}_i$). 

In order to specify constraints defining $\delta_{ij}$, we need to distinguish the case where criterion $i$ is a single-peaked or a single-valley criterion. In the first case, the single-peaked criterion is transformed into a cost criterion and the following constraints hold : 
\begin{subequations}
\begin{align}
& \delta_{ij}= 1 \Longleftrightarrow  |\alpha^j_i| \leq b_{i}  \Longrightarrow  M(\delta_{ij} - 1) \leq b_{i} - (\alpha^{j+}_i + \alpha^{j-}_i) \label{eq:sp1} \\ 
& \delta_{ij}= 0 \Longleftrightarrow |\alpha^j_i| > b_{i} \Longrightarrow   b_{i} - (\alpha^{j+}_i + \alpha^{j-}_i) < M \, \delta_{ij} \label{eq:sp2} \\
& \delta_{ij} \in \{0,1\} \label{eq:delta}
\end{align}
\end{subequations}
In the second case, the single-valley criterion is transformed conversely into a gain criterion as follows : 
\begin{subequations}
\begin{align}
& \delta_{ij}= 1 \Longleftrightarrow  |\alpha^j_i| \geq b_{i}  \Longrightarrow M(\delta_{ij} - 1) \leq (\alpha^{j+}_i + \alpha^{j-}_i) - b_{i} \label{eq:sv1} \\
& \delta_{ij}= 0 \Longleftrightarrow |\alpha^j_i|< b_{i}  \Longrightarrow (\alpha^{j+}_i + \alpha^{j-}_i) - b_{i} < M \, \delta_{ij} \label{eq:sv2} \\
& \delta_{ij} \in \{0,1\} \tag{\ref{eq:delta}}
\end{align}
\end{subequations}
In order to jointly consider both cases (\ref{eq:sp1})-(\ref{eq:sp2}) and (\ref{eq:sv1})-(\ref{eq:sv2}) in the MIP, we introduce a binary variable $\sigma_i, \; i \in {\mathcal N}$ which indicates whether criterion $i$ is a single-peaked ($\sigma_i=1$) or single-valley criterion ($\sigma_i=0$). When $\sigma_i=1$, the constraints (\ref{eq:sp-sv3}) and (\ref{eq:sp-sv4}) concerning the single-peaked criteria hold while the constraints (\ref{eq:sp-sv1}) and (\ref{eq:sp-sv2}) for single-valley criteria are relaxed, and conversely when  $\sigma_i=0$. 

\begin{subequations}
\begin{align}
& - M \, \sigma_i + M(\delta_{ij} - 1) \leq \alpha^{j+}_i + \alpha^{j-}_i - b_{i} \label{eq:sp-sv1} \\
& \alpha^{j+}_i + \alpha^{j-}_i - b_{i} < M \, \delta_{ij} + M \, \sigma_i \label{eq:sp-sv2} \\
& M.(\sigma_i-1) + M(\delta_{ij} - 1) \leq b_{i} - \alpha^{j+}_i - \alpha^{j-}_i \label{eq:sp-sv3} \\
& b_{i} - \alpha^{j+}_i - \alpha^{j-}_i < M \, \delta_{ij} + M \, (1-\sigma_i) \label{eq:sp-sv4} \\
& \delta_{ij} \in \{0,1\} \tag{\ref{eq:delta}} \\
& \sigma_{i} \in \{0,1\} \label{eq:sigma}
\end{align}
\end{subequations}

Lastly, in order to restrain the bounds of the single-peaked/single-valley interval within $[min_i,max_i]$, we add the 2 following constraints :
\begin{subequations}
\begin{align}
b^{\bot}_i - b_{i} \geq min_i , \;\;\; b^{\bot}_i + b_{i} \leq max_i \label{eq:prof_limit}
\end{align}
\end{subequations}

%----------------------------
\subsection{Variables and constraints related to weights}
\label{subsec:dec_var_weights}
%----------------------------

As in~\cite{leroy2011learning}, we define the continuous variables $c_{ij}, \; i \in {\mathcal N}, \; j \in J^{*}$ such that $\delta_{ij} = 0 \Leftrightarrow c_{ij}=0$ and $\delta_{ij} = 1 \Leftrightarrow c_{ij}=w_i$, where $w_i\ge 0$ represents the weight of criterion $i$ with the  normalization constraint: $\sum_{\forall i \in \mathcal{N}} w_i = 1$. To ensure the correct definition of $c_{ij}$, we impose:

\begin{subequations}
\begin{align}
& c_{ij} \leq \delta_{ij} \label{eq:delta_weigths3} \\
& \delta_{ij} - 1 + w_i \leq c_{ij} \label{eq:delta_weigths4} \\
& c_{ij} \leq  w_{i} \label{eq:delta_weigths5} \\
& 0 \leq  c_{ij} \label{eq:delta_weigths6}
\end{align}
\end{subequations}

%----------------------------
\subsection{Variables and constraints related to the assignment examples}
%----------------------------

So as to check whether assignment examples are correctly restored by the MR-Sort rule, we define binary variables $\gamma_j\in \{0,1\}, \; j \in J^{*}$ equal to 1 when the alternative $a^j$ is correctly assigned, 0 otherwise. The constraints bellow guaranties the correct definition of  $\gamma_j$ (where $\lambda \in [0.5, 1]$ represents the MR-Sort majority threshold).
\begin{subequations}
\begin{align}
& \textstyle\sum_{i \in \mathcal{N}} c_{ij}  \geq \lambda + M (\gamma_j - 1), \forall j \in J^{*2}  \label{eq:mrsort_rule1} \\
& \textstyle\sum_{i \in \mathcal{N}} c_{ij}  < \lambda - M (\gamma_j - 1), \forall j \in J^{*1} \label{eq:mrsort_rule2}
\end{align}
\end{subequations}

%----------------------------
\subsection{Objective function}
\label{subsec:dec_var_obj_func}
%----------------------------

The objective for the $\textit{Inv-MR-Sort}$ problem is to identify the MR-Sort model which best matches the learning set. Therefore, in order to maximize the number of correctly restored assignment examples, the objective function can be formulated as: $Max \sum_{j \in J^*} \gamma_j$. Finally, the MIP formulation for the $\textit{Inv-MR-Sort}$ problem with single-peaked and single valley criteria is given below (where  $M$ is an arbitrary large positive value, and $\varepsilon$ an arbitrary small positive value). Table~\ref{tab:dec-var-description} synthesizes the variables involved in this mathematical program.

\begin{subequations}
\begin{align}
    & \max \sum_{j \in J^*} \gamma_j  & \label{eq:obj} \\
    & \textstyle\sum_{i \in \mathcal{N}} c_{ij} \geq \lambda+ M(\gamma_j-1) & \forall j \in J^{*2} \tag{\ref{eq:mrsort_rule1}} \\ 
    & \textstyle\sum_{i \in \mathcal{N}} c_{ij}+\varepsilon \leq \lambda- M(\gamma_j-1) &  \forall j \in J^{*1} \tag{\ref{eq:mrsort_rule2}} \\ 
    & \textstyle\sum_{i \in \mathcal{N}} w_i = 1         \\
    & c_{ij} \leq \delta_{ij}  &                     \forall j \in J^*, \forall i \in \mathcal{N} \tag{\ref{eq:delta_weigths3}} \\ 
    & c_{ij} \geq \delta_{ij} - 1 + w_{i} &           \forall j \in J^*, \forall i \in \mathcal{N} \tag{\ref{eq:delta_weigths4}} \\
    & c_{ij} \leq  w_{i} &                              \forall j \in J^*, \forall i \in \mathcal{N} \tag{\ref{eq:delta_weigths5}}  \\ 
    & b^{\bot}_i- a^j_i = \alpha^{j+}_i - \alpha^{j-}_i &    \forall j \in J^*, \forall i \in \mathcal{N} \tag{\ref{eq:abs-beta2}} \\ 
    & \alpha^{j+}_i \leq  \beta^j_i M &                 \forall j \in J^*, \forall i \in \mathcal{N} \tag{\ref{eq:abs-beta3}} \\ 
    & \alpha^{j-}_i \leq (1-\beta^j_i) M &              \forall j \in J^*, \forall i \in \mathcal{N} \tag{\ref{eq:abs-beta4}} \\ 
    & - M.\sigma_i + M(\delta_{ij} - 1) \leq \alpha^{j+}_i + \alpha^{j-}_i - b_{i} &  \forall j \in J^*, \forall i \in \mathcal{N} \tag{\ref{eq:sp-sv1}} \\ 
    & \alpha^{j+}_i + \alpha^{j-}_i - b_{i} + \varepsilon \leq M.\delta_{ij} + M.\sigma_i &  \forall j \in J^*, \forall i \in \mathcal{N} \tag{\ref{eq:sp-sv2}} \\
    & M.(\sigma_i-1) + M(\delta_{ij} - 1) \leq b_{i} - \alpha^{j+}_i - \alpha^{j-}_i &  \forall j \in J^*, \forall i \in \mathcal{N} \tag{\ref{eq:sp-sv3}} \\ 
    & b_{i} - \alpha^{j+}_i - \alpha^{j-}_i + \varepsilon \leq M.\delta_{ij} + M.(1-\sigma_i) &  \forall j \in J^*, \forall i \in \mathcal{N} \tag{\ref{eq:sp-sv4}} \\ 
    & b^{\bot}_i - b_{i} \geq min_i,\; b^{\bot}_i + b_{i} \leq max_i &                            \forall i \in \mathcal{N} \tag{\ref{eq:prof_limit}} \\
     & c_{ij} \in [0,1], \delta_{ij} \in \{0,1\} &                                \forall j \in J^*, \forall i \in \mathcal{N} \label{eq:mip2} \\
    & \alpha^{j+}_i,\alpha^{j-}_i  \in \mathbb{R}^+ &                    \forall j \in J^*, \forall i \in \mathcal{N} \\ 
    & \beta^{j}_i \in [0,1] &                           \forall j \in J^*, \forall i \in \mathcal{N} \\ 
    & b_i \in \mathbb{R}, w_{i} \in [0,1], b^{\bot}_i \in \mathbb{R}, \sigma_i \in \{0,1\} &                              \forall i \in \mathcal{N} \\
    & \gamma_{j} \in \{0,1\} &                          \forall j \in J^*  \\
    & \lambda \in [0.5,1]                               \label{eq:mip10}
\end{align}
\end{subequations}

\begin{table}[ht!]
\centering
\begin{tabular}{cccl}
\hline Variable $\;$ & Domain $\;$& Number of variables $\;$& Definition \\ \hline 
$\alpha^{j+}_i$ &  $\mathbb{R}^+$ & $n\times|A^\star|$ & first component of the absolute value $|a^{j}_i-b_i^\bot|$ \\
$\alpha^{j-}_i$ & $\mathbb{R}^+$ & $n\times|A^\star|$ & second component of the absolute value of $|a^{j}_i-b_i^\bot|$ \\
$\beta^j_i$     & $\{0,1\}$ & $n\times|A^\star|$ & binary variable indicating the sign of $a^{j}_i-b_i^\bot$  \\

$\sigma_i$      & \{0,1\} & $n$ &  $\sigma_i=1$ if criterion $i$ is single-peaked, $\sigma_i=0$ if $i$ is single-valley \\ 

$\gamma_{j}$    & \{0,1\} & $|A^\star|$  & $\gamma_j=1$ if alternative $a^j$ is correctly assigned by the model, $\gamma_j=1$ if not  \\

$\delta_{ij}$   & \{0,1\} & $n\times|A^\star|$ & $\delta_{ij}=1$ if $a^j_i \in {\mathcal A}_i$, $\delta_{ij}=0$ if $a^j_i \notin {\mathcal A}_i$  \\

$c_{ij}$        & [0,1] & $n\times|A^\star|$ & $c_{ij}=1$ if $a^j_i \in {\mathcal A}_i$ (i.e, if $\delta_{ij}=1$), $c_{ij}=0$ if $a^j_i \notin {\mathcal A}_i$ (i.e, if $\delta_{ij}=0$)  \\
$b^{\bot}_i$    & $\mathbb{R}$ & $n$  & middle of the interval $[\underline{b}_i, \overline{b}_i]$\\
$b_{i}$         & $\mathbb{R}$ & $n$  & value of half the width of the interval $[\underline{b}_i, \overline{b}_i]$ on criterion $i$ \\
$w_{i}$         & [0,1] & $n$  & weight of criterion $i$ \\
$\lambda$       & [0,1] & 1 &  majority threshold \\
\hline
\end{tabular}
\caption{Description of decision variables}
\label{tab:dec-var-description}
\end{table}

%----------------------------
\subsection{Interpretation of the optimal solution}
\label{subsec:retriving_parameters}
%----------------------------

Once the above mathematical program is solved, it is necessary to derive the corresponding MR-Sort model:  
\begin{itemize}
    \item the nature of each criterion (either cost, gain, single-peaked, or single-valley criterion),
    \item the weights attached to criteria $w_i$, and associated majority level $\lambda$,
    \item the frontier between category $C^1$ and $C^2$, i.e., the value $b_i$ if criterion $i$ is a cost or a gain criterion, and the interval $[\underline{b}_i, \overline{b}_i]$  if criterion $i$ is a single-peaked or single-valley criterion.
\end{itemize}

\begin{figure}[h!]
    \centering
    \includegraphics[scale=1.1]{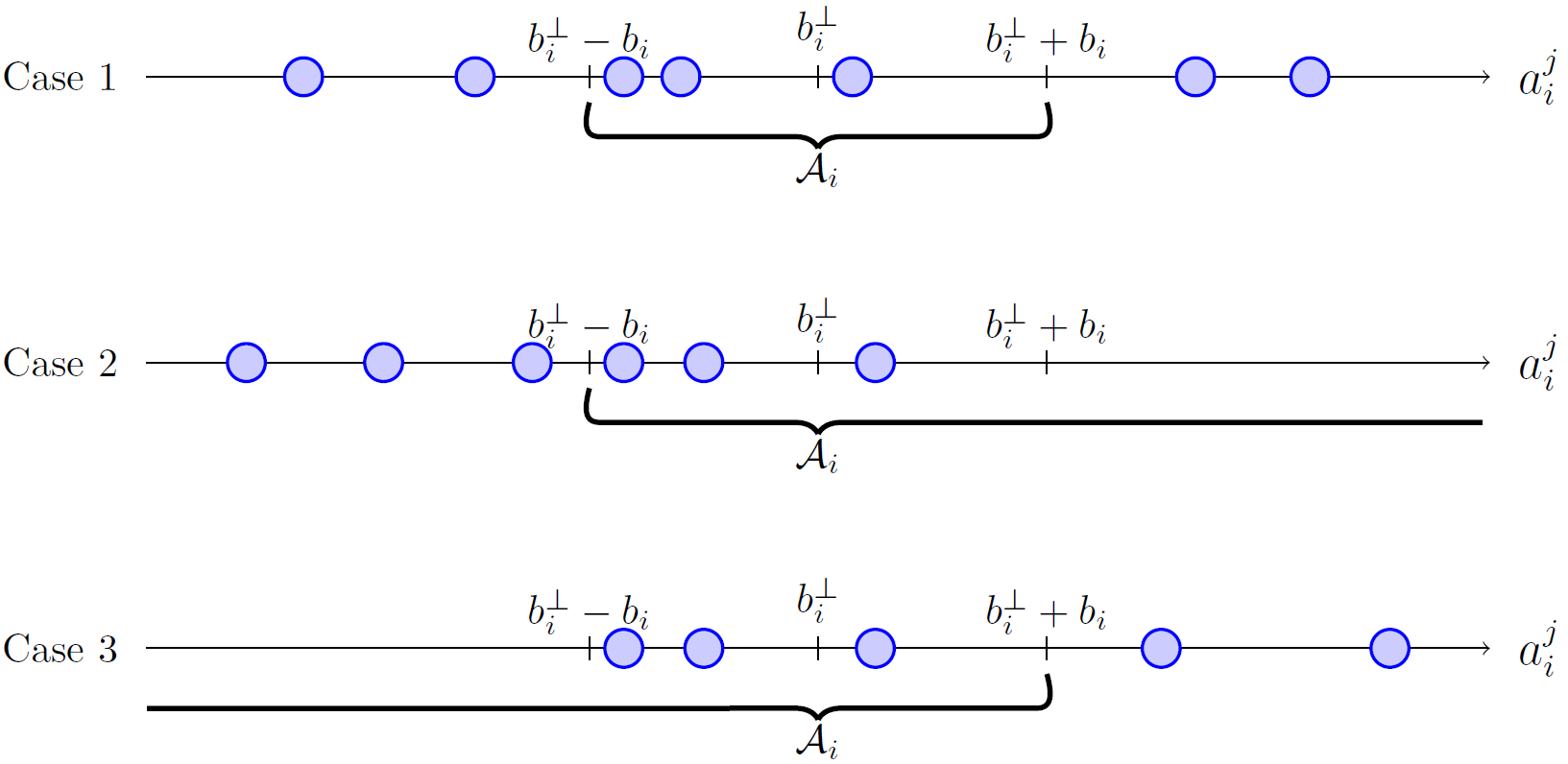}
    \caption{Three cases for single-peaked criteria}
    \label{casesSP}
\end{figure}

\begin{figure}[h!]
    \centering
    \includegraphics[scale=1.1]{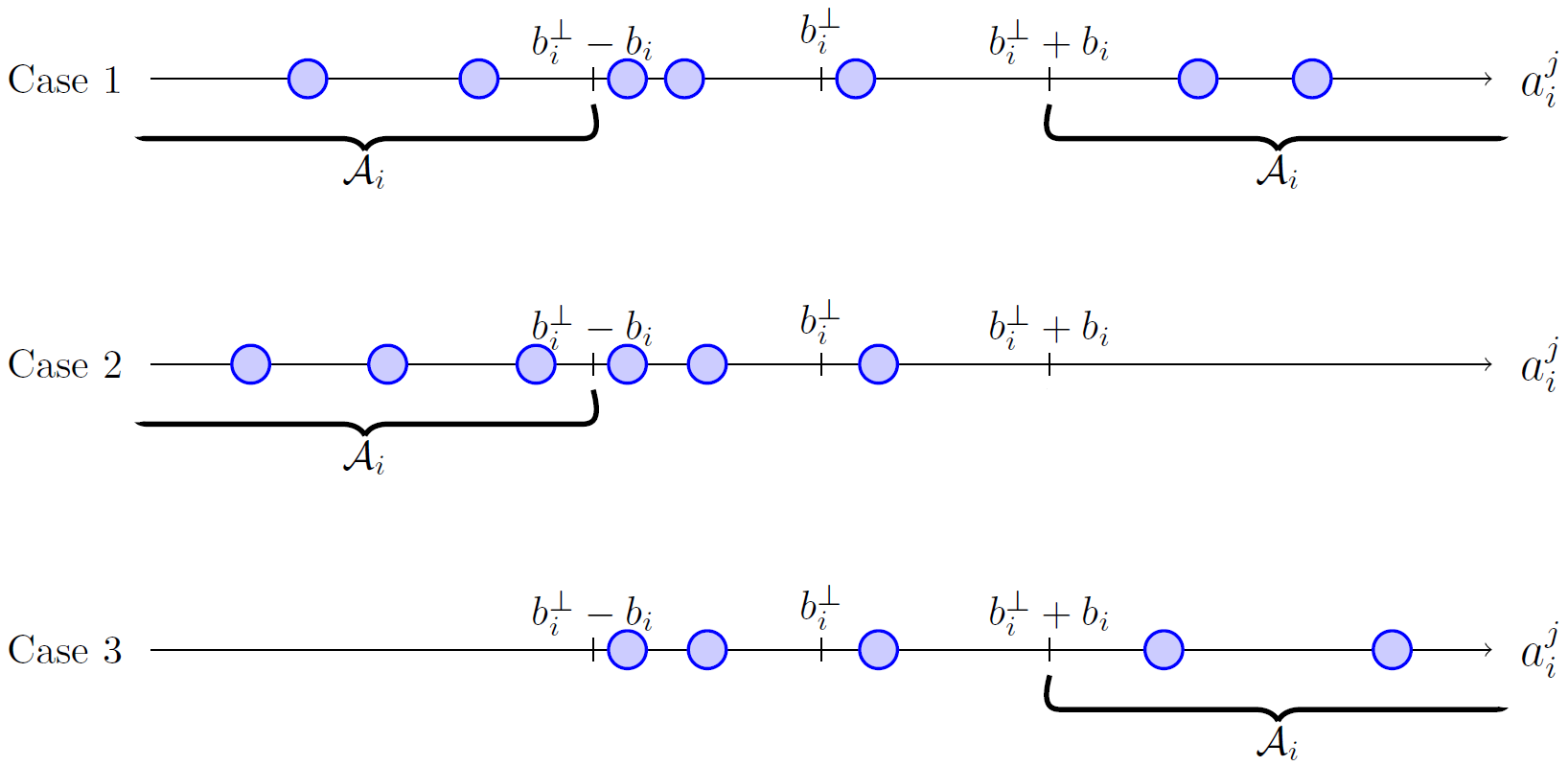}
    \caption{Three cases for single-valley criteria}
    \label{casesSV}
\end{figure}

%\newpage
Criteria weights $w_i$, and associated majority level $\lambda$ are directly obtained from the corresponding variables in the optimal solution. The preference directions and criteria limit profiles are deduced as  follows:
\begin{itemize}
    \item Case $\sigma_i=1$ (i.e. criterion $i$ is represented as a single-peaked criterion in the optimal solution):%, see Figure~\ref{casesSP}):
    \begin{itemize}
        \item if $b^{\bot}_i-b_i \leq min_{j \in J^*}\{a_i^j\}$, then criterion $i$ is a cost criterion, and the maximal approved value on criterion $i$ is $b^{\bot}_i+b_i$, i.e. ${\mathcal A}_i=]-\infty,b^{\bot}_i+b_i]$, see Figure~\ref{casesSP}~case~3,
        \item if $b^{\bot}_i+b_i \geq max_{j \in J^*}\{a_i^j\}$, then criterion $i$ is a gain criterion, and the minimal approved value on criterion $i$ is $b^{\bot}_i-b_i$, i.e. ${\mathcal A}_i=[b^{\bot}_i-b_i, \infty[$, see Figure~\ref{casesSP})~case~2,
        \item otherwise, $i$ is a single-peaked criterion, and ${\mathcal A}_i = [b^{\bot}_i-b_i,b^{\bot}_i+b_i]$, see Figure~\ref{casesSP}~case~1% \subset [min_i,max_i]$
    \end{itemize}
    \vspace*{1mm}
    \item Case $\sigma_i=0$ (i.e. criterion $i$ is represented as a single-valley criterion in the optimal solution):%, see Figure~\ref{casesSV}):
    \begin{itemize}
        \item if $b^{\bot}_i-b_i < min_{j \in J*}\{a_i^j\}$, then criterion $i$ is a gain criterion, and the minimal approved value on criterion $i$ is $b^{\bot}_i+b_i$, i.e. ${\mathcal A}_i=[b^{\bot}_i+b_i,\infty[$, see Figure~\ref{casesSV}~case~3,
        \item if $b^{\bot}_i+b_i > max_{j \in J^*}\{a_i^j\}$, then criterion $i$ is a cost criterion, and the maximal approved value on criterion $i$ is $b^{\bot}_i-b_i$, i.e. ${\mathcal A}_i=[-\infty,b^{\bot}_i-b_i]$, see Figure~\ref{casesSV}~case~2,
        \item otherwise, $i$ is a single-valley criterion, and ${\mathcal A}_i = [-\infty,b^{\bot}_i-b_i]\cup [b^{\bot}_i+b_i,\infty[$, see Figure~\ref{casesSV}~case~1.% \subset [min_i,max_i]$. 
    \end{itemize}
\end{itemize}

%----------------------------
\subsection{Extension to more than 2 categories}
\label{subsec:more-than-2-categories}
%----------------------------
Our framework can be extended to more than two categories, at the cost of adding supplementary variables and constraints to the mathematical program. So as to extend to $p$ categories ($p>2$), sets of approved values $\mathcal{A}_i^h \subseteq X_i$ on criterion $i$ ($i \in \mathcal{N}$) should be defined with respect to a category level $h$ ($h=1, 2, \hdots, p-1$), and should be embedded such that $\mathcal{A}^{p-1}_i \subseteq \mathcal{A}^{p-2}_i \subseteq ... \subseteq \mathcal{A}^{1}_i$. 

In the MIP formulation, the variables $\delta_{ij}$, $c_{ij}$, $\alpha_{i}^{j+}$, $\alpha_{i}^{j-}$, $\beta_i^{j}$, $b_{i}^{}$, and $b_{i}^{\bot}$ should be indexed with a category level $h=1..p-1$, and become $\delta_{ij}^h$, $c_{ij}^h$, $\alpha_{i}^{jh+}$, $\alpha_{i}^{jh-}$, $\beta_i^{jh}$, $b_{i}^{h}$, and $b_{i}^{h\bot}$, respectively.
Constraints (\ref{eq:mrsort_rule1}) and (\ref{eq:mrsort_rule2}) relative to the assignment examples should be replaced by the following ones:

\begin{itemize}

    \item[$\bullet$] $\sum_{i \in \mathcal{N}} c_{ij}^{p-1} \geq \lambda+ M(\gamma_j-1),\; \forall a_j\in C^p$
    
    \item[$\bullet$] $\sum_{i \in \mathcal{N}} c_{ij}^1+\varepsilon \leq \lambda- M(\gamma_j-1),\;   \forall a_j\in C^1$
    
    \item[$\bullet$] $\sum_{i \in \mathcal{N}} c_{ij}^{h-1} \geq \lambda+ M(\gamma_j-1),\;   \forall a_j\in C^h \subset [C^2, C^{p-1}]$
    
    \item[$\bullet$] $\sum_{i \in \mathcal{N}} c_{ij}^h+\varepsilon \leq \lambda- M(\gamma_j-1),\;   \forall a_j\in C^h \subset [C^2, C^{p-1}]$
\end{itemize}

Lastly, constraints on $b_{i}^{h}$, and $b_{i}^{h\bot}$ should be imposed so as to guarantee that the approved sets are embedded such that $\mathcal{A}^{p-1}_i \subseteq \mathcal{A}^{p-2}_i \subseteq ... \subseteq \mathcal{A}^{1}_i$, i.e, $[b_i^{p-1\bot}-b_i^{p-1}, b_i^{p-1\bot}+b_i^{p-1}] \subseteq [b_i^{p-2\bot}-b_i^{p-2}, b_i^{p-2\bot}+b_i^{p-2}] \subseteq \ldots \subseteq [b_i^{1\bot}-b_i^{1}, b_i^{1\bot}+b_i^{1}]$.

%===============================
\section{Experiments, results and discussion}
\label{sec:experiments_results}
%===============================

In this section, we report numerical experiments to empirically study how the proposed algorithm behaves in terms of computing time, ability to generalize, and ability to restore an MR-Sort model with the correct preference direction (gain, cost, single-peaked, or single-valley).  The experimental study involves artificially generated datasets and an ex-post analysis of a real-world case study.

\subsection{Tests on generated datasets}

\subsubsection{Experimental design}

\noindent
Assuming a generated MR-Sort model $\mathcal{M}^0$ perfectly representing the Decision Maker preferences, we first randomly generate n-tuples of values considered as alternatives (each tuple corresponding to $n$ criteria evaluations). Then we simulate the assignments of these alternatives following the model $\mathcal{M}^0$ and obtain assignment examples which constitute the learning set $L$, used as input to our MIP algorithm. Alternatives are generated in such a manner as to  obtain a balanced dataset (i.e. equal number of assignments in each category). The Inv-MR-Sort problem is then solved using the proposed algorithm and, as a result, generating a learned model noted $\mathcal{M}^\prime$.

\newblock

\noindent\textbf{Generation of instances and model parameters.} 
We consider a learning set of 200 assignment examples. 
A vector of performance values of alternatives is drawn in an independent and identically distributed manner, such that the performance values are contained in the unit interval discretized by tenths. We then randomly generate profile values (either $b_i$, or $\underline{b}_i$ and $\overline{b}_i$) for each criterion; these values are also chosen with the unit interval discretized by tenths.

In order to draw uniformly distributed  weight vectors~\cite{BUTLER1997531}, we uniformly generate $|\mathcal{N}|-1$ random values in $[0,1]$ sorted in ascending order. We then prepend 0 and append 1 to this set of values obtaining a sorted set of  $|\mathcal{N}|+1$ values. Finally, we compute the difference between each successive pair of values resulting in a set of $|\mathcal{N}|$ weights such that their sum is equal to 1. We randomly draw $\lambda$ in [0,1].

In order to assess the ability of the algorithm to restore preference directions, we randomly assign to $q$ criteria over $n$, a preference direction among the four (gain, cost, single-peaked and single-valley). 
Hereby, the preference direction of these criteria are assumed to be unknown in the learning set, meanwhile the remaining $n-q$ criteria are considered as gain criteria. 

\newblock

\noindent\textbf{Performance metrics and tests parameters.}
To study the performance of the proposed algorithm, we consider three main metrics.
\begin{itemize}
\item \emph{Computing time}: the time (CPU) necessary to solve the MIP algorithm.

\item \emph{Restoration rate of assignment examples}: as our MIP algorithm is an exact method, it is expected that the entire learning set $L$ will be restored by $\mathcal{M}^\prime$. Therefore, we assess the restoration performance on a test set which is run  through  $\mathcal{M}^0$ and $\mathcal{M}^\prime$. This test set comprises randomly generated alternatives that were not used in the learning set; that is, assignment examples that the algorithm has never seen. This allows us to assess the restoration rate (also called classification accuracy in generalization or $CAg$), that is the ratio between the number of alternatives identically assigned in categories by both $\mathcal{M}^0$ and $\mathcal{M}^\prime$, and the number of alternatives. 

\item \emph{Preference direction restoration rate (PDR)}: considering the set of criteria where the preference direction is unknown, PDR is defined as the ratio between the number of criteria where the preference direction has been correctly restored and the cardinality of this set. 
\end{itemize}

In order to account for the statistical distribution of all the randomly selected values, we independently select 100 different learning sets, each one associated to a randomly generated $\mathcal{M}^0$ MR-Sort model. We then run 100 independent experiments and aggregate the results.

In our experiments, the number of criteria $n$ are considered in  $\{4,5,6,7,8,9\}$, $q$ being the number of criteria with unknown preference directions in $\{0,1,2,3,4\}$, and the number of categories is set to 2. The test set is composed of 10000 randomly generated alternatives.

We executed experiments on a server endowed with an Intel Xeon\footnote{Intel, and Intel Xeon are trademarks or registered trademarks of Intel Corporation or its subsidiaries in the United States and other countries.} Gold 6248 CPU @ 2.50GHz, 80 cores and 384 GB RAM. CPLEX 20.1~\cite{ibmurl}  was used for the MIP resolution. In order to ensure the uniformity in tests in terms of CPU, we limit its use to 10 threads per test. In the following, we use CPLEX with the same settings except for this section, we adopt a time out at 1 hour.

%\newblock

\subsubsection{Results}
%\newblock\\

\noindent 
In the following, we present the results of the randomly generated tests.

\begin{table}[H]
\def\arraystretch{2}
\centering
\begin{tabular}{|c|cccccc|}
\cline{1-7}
$\#$ unknown & \multicolumn{6}{c|}{Number of criteria ($n$)} \\ 
 \cline{2-7}
direction ($q$)   & $4$ & $5$ & $6$ & $7$ & $8$ & $9$  \\ \hline
$0$ & 0.34 (\textbf{100}) & 0.56 (\textbf{100}) & 0.84 (\textbf{100}) & 2.38 (\textbf{100}) & 2.61 (\textbf{100}) & 3.37 (\textbf{100})  \\
$1$ & 1.51 (\textbf{100}) & 3.23 (\textbf{100}) & 4.53 (\textbf{100}) & 7.22 (\textbf{100}) & 19.15 (\textbf{100}) & 18.68 (\textbf{91}) \\
$2$ & 6.12 (\textbf{100}) & 12.97 (\textbf{100}) & 30.06 (\textbf{94}) & 54.38 (\textbf{90}) & 43.19 (\textbf{72}) & 58.03 (\textbf{59}) \\
$3$ & 37.48 (\textbf{95}) & 76.68 (\textbf{89}) & 72.46 (\textbf{80}) & 76.29 (\textbf{54}) & 59.32 (\textbf{47}) & 25.91 (\textbf{31}) \\
$4$ & 96.34 (\textbf{59}) & 129.49 (\textbf{57}) & 61.01 (\textbf{52}) & 108.25 (\textbf{42}) & 22.17 (\textbf{27}) & 23.63 (\textbf{28}) \\ \hline
\end{tabular}
\caption{Median CPU Time (sec.) of instances solved in 1h, and number of terminated instances in parentheses,  with 4 to 9 criteria ($n$), and 0 to 4 criteria with unknown pref. direction ($q$)}
\label{tab:table-cpu-tests}
\end{table}

\noindent\textbf{Computing time performance.}
Table~\ref{tab:table-cpu-tests} presents the median CPU time of the terminated instances (timeout fixed to 1 hour). The execution time increases with the number of criteria and the number of criteria with unknown preference direction up to $n=7$ and $q=2$. Beyond this limit, the execution time fluctuates as it is influenced by the number of terminated instances which are taken into account in the median time computation.   

Additionally, Table~\ref{tab:table-cpu-tests} shows the percentage of instances that terminated within the time limit, set to 1 hour. 

Unsurprisingly, the number of terminated instances decreases with both the number of criteria and the number of criteria with unknown preference directional. In particular, the rate jumps from 95\% with 4 criteria to 35\% with 9 criteria in the model when $q=3$.

\newblock

\noindent\textbf{Restoration rate of the test set.}
Regarding the classification accuracy (CAg) of the learned models (involving 4 to 9 criteria in the model and 0 to 4 criteria with unknown preference direction), the performance values are globally comprised between 0.9 and 0.95 with 0.93 on average.
We do not notice a significant trend over both the number of criteria and the number of criteria with unknown preference directions. 
However, the figures reflect the performance of only terminated instances. Therefore, the CAg rate could possibly degrade when taking into account executions above the timeout, assuming these are the most difficult instances to learn. 

\begin{figure*}[h!]
\centering
  \includegraphics[scale=0.6]{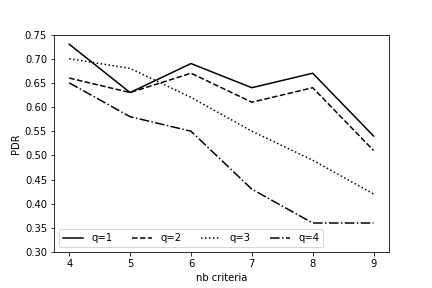}
  \caption{Preference direction restoration rate (PDR) considering 1 to 4 criteria with unknown preference direction ($q$) (average performance over terminated instances)}
  \label{plot:plot-pdr}
\end{figure*}

\begin{table}[H]
\def\arraystretch{2}
\centering
\begin{tabular}{c|c|c|c|}
\cline{2-4}
 & $w_1 \le \frac{1}{2n}$ & $ \frac{1}{2n} < w_1 < \frac{2}{n}$ & $ w_1 \ge \frac{2}{n}$ \\ \hline
\multicolumn{1}{|c|}{PDR} & 0.44 & 0.74 & 0.78 \\ \hline
\end{tabular}
\caption{PDR (averaged over $n$) according to the range of weight of criterion $c_1$}
\label{tab:table-inflluence-w}
\end{table}

\noindent\textbf{Preference direction restoration rate.}
Figure~\ref{plot:plot-pdr} illustrates the evolution of the preference direction restoration rate (PDR). 

Globally, the PDR falls with the increase of number of criteria in the model. In addition, this indicator degrades moderately with the number of criteria with unknown preference directions with respectively 55\% and 35\% for $q=1$ and $q=4$ considering 9 criteria in the model.

The results illustrated in the Table~\ref{tab:table-inflluence-w} give more insights into the behaviour of the algorithm regarding PDR. We consider instances involving one criterion with unknown preference direction, $q=1$ (it corresponds to criterion 1).

 We analyze the impact of this criterion ($w_1$) on the preference direction rate of restoration rate. The PDR rate is averaged over the number of criteria in the model ($n \in \{4,..,9\}$) and distributed over three intervals:
 
$[0,\frac{1}{2n}]$ ,$]\frac{1}{2n},\frac{2}{n}[$, $[\frac{2}{n},1]$ which can be interpreted as three levels of importance of $w_1$ (respectively low level, medium level, and high level).
As expected, the average PDR rises with the importance of $w_1$; we have 44\% of PDR for low level of importance, whereas 74\% and 78\% correspond respectively to a medium and high level of the importance of $w_1$. It appears that the MIP has more difficulty in correctly detecting the preference direction of a criterion when this criterion has a low importance.

\subsubsection{Discussion}
%\newblock\\

The experiments carried out on randomly generated instances give us the following insights.\\[-3mm]

Although exact methods are typically computationally intensive, the computation time is relatively affordable for medium-sized models (less than 3 minutes for 200 alternatives in the learning set and up to $n=9$ and $q=4$ in the model when the timeout is set to 1 hour). Moreover, the computation time could be reduced as our experiments were performed with a limited number of threads set to 10.

The algorithm is able to accurately restore new assignment examples based on the learned models (0.93 on average up to 9 criteria) and remains relatively efficient with the number of criteria with unknown preference directions. Extended experiments should be done without the limit of time to accurately predict the generalization restoration rate with the increase of parameters $n$ and $q$.                            

Our algorithm struggles to restores preference directions when the number of criteria grows while keeping the learning set constant.
The PDR rate also decreases with the increase of the number of criteria with unknown preference direction in the model with similar learning set sizes (but still greater than the random choice that is  25\%). For more insight, it would be instructive to discover the algorithm's behaviour in terms of the PDR for non-terminated instances.

Finally, the restoration rate of criteria preference direction is correlated with the importance of such criteria in the model. It appears that the preference direction of criteria with an importance below  $\frac{1}{2n}$ is the most difficult to restore. These results are valid with a learning set of fixed size (200). It would be interesting to further investigate experimentally whether larger learning sets would make it possible to accurately learn the direction of preference.

\subsection{Tests on a real-world data: the ASA dataset}

The ASA\footnote{ASA stands for ``\emph{American Society of Anesthesiologists}''.} dataset~\cite{Lazouni_2013} constitutes a list of 898 patients evaluated on 14 medical indicators (see Table~\ref{tab:asa-original}) enabling assignment of patients into 4 ordered categories (ASA1, ASA2, ASA3, ASA4). These categories correspond to 4 different scores that indicate the patient health. Based on the score obtained for a given patient, anesthesiologists decide whether or not to admit such a patient to surgery.
The relevance of the dataset for our tests relies on the presence of a criterion with single-peaked preference, which is ``Blood glucose level" (i.e. glycemia). For the sake of practicality,  for our experiments we restrain the ASA dataset to the 8 most relevant criteria. They are in bold in Table~\ref{tab:asa-original}

\begin{table}[h]
\centering
\begin{tabular}{lcc}
\hline Attribute & Domain (Unit) & $\;$Direction \\
\hline \textbf{Age} & {$[0-105]$ (year) } & min. \\
\textbf{Diabetic} & \{0,1\} & min. \\
\textbf{Hypertension} & \{0,1\} & min. \\
\textbf{Respiratory failure} & \{0,1\} & min. \\
Heart failure & \{0,1\} & min. \\
Heart rate & {$[55-123]$ (bpm) } & SP \\
Heart rate steadiness & \{0,1\} & max. \\
\textbf{Pacemaker} & \{0,1\} & min. \\
Atrioventricular block & \{0,1\} & min. \\
Left ventricular hypertrophy & \{0,1\} & $\min .$ \\
Oxygen saturation & {$[43-100](\%)$} & max. \\
\textbf{Blood glucose level (glycemia)} $\;$ & {$[0.5-3.8](\mathrm{g} / \mathrm{l})$} & SP \\
\textbf{Systolic blood pressure} & {$[9-20.5](\mathrm{cm} \mathrm{Hg})$} $\;$ & min. \\
\textbf{Diastolic blood pressure} & {$[5-13](\mathrm{cm} \mathrm{Hg})$} & min. \\
\hline
\end{tabular}
\caption{Original criteria in the ASA dataset}
\label{tab:asa-original}
\end{table}

To have two categories, we first divide the dataset into two parts: \emph{category 2} representing patients in categories ASA1 and ASA2 (67\% of the population) and \emph{category 1} representing those in categories ASA3 and ASA4 (33\% of the population).\\[-3mm]

In the following, we illustrate how to learn the model parameters and the preference type (gain, cost, single-peaked (SP), single-valley) of the criterion ``Glycemia'' with three different sets of assignment examples chosen from the original set of 898 patients. In this medical application, we suppose that the ``Glycemia'' criterion type is unknown and expect to ``discover'' a single-peaked criterion.

For each experiment we report the number of distinct performances considered per criterion.

\paragraph*{\textbf{First Dataset:}} we initially consider the whole original dataset with all 898 assignment examples in the learning set as input to our MIP algorithm. From this first dataset, we infer the type (gain, cost, single-peaked, single-valley) of criterion Glycemia and the  MR-Sort parameters. \\[-3mm]

\begin{figure}[ht!]
  \centering
  \includegraphics[scale=0.7]{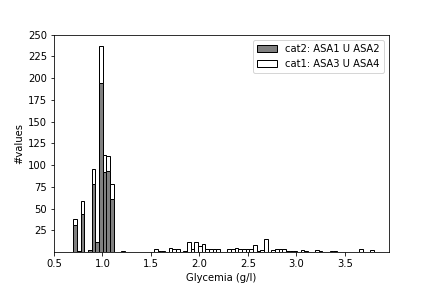}
  \caption{Distribution of patients' glycemia in the first dataset 
  }
  \label{fig:hist-phase1}
\end{figure}

The inferred model given in Table~\ref{tab:asa-phase1} is computed in $40h33mn$ execution time. The obtained model restores $CA=99.4\%$ of the assignment examples in the learning set. However, in the inferred model, the glycemia criterion is detected as a cost criterion to be minimized (whereas we expect it to be inferred as single-peaked). Note that the inferred value for the limit profile on the glycemia criterion (1.18 g/l) makes it possible to distinguish patients with hyperglycemia from the others, but does not distinguishes hypoglycemia from normal glycemia (normal glycemia corresponds to [0.9,1.2]). This is due to the distribution of the glycemia values over the patients shown in Figure~\ref{fig:hist-phase1}. In this distribution we observe that all patients with glycemia above 1.2g/l (hyperglycemia) are assigned to Category 1. However, some patients with normal glycemia [0.9,1.2] are also assigned to Category 1, and some patients with glycemia equal to 0.8 g/l or below (hypoglycemia) are assigned to Category 2.

In the following, we check if it is possible to restore the ``correct'' preference direction (i.e. single-peaked) with a subset of carefully selected patients. To do so we will remove the patients with normal glycemia ($[0.9,1.2]$ g/l) assigned to category 1.

\begin{table}[ht!]
\centering
\begin{tabular}{|c|ccc|>{\centering\arraybackslash}p{1.2cm}>{\centering\arraybackslash}p{1.2cm}>{\centering\arraybackslash}p{1.4cm}|} \hline 
& \multicolumn{3}{|c}{\textbf{\begin{tabular}[c]{@{}c@{}} Instance settings \end{tabular}}} & 
\multicolumn{3}{|c|}{\textbf{\begin{tabular}[c]{@{}c@{}} Model parameters learned \end{tabular}}}  \\
Attributes & $\#values$ & Direction & pref. dir.  & $b_i$ & $w_i$ & pref. dir. \\ \hline 
Age & 103 (origin) & min. & known & 72.9 & 0.01 & $\_$   \\
Diabetic &  2 (origin) & min. & known & 0.99 & 0 & $\_$  \\
Hypertension &  2 (origin) & min. & known & 0 & 0.01 & $\_$  \\
Respiratory F & 2 (origin) & min. & known & 0.99 & 0.88 & $\_$  \\
Pacemaker & 2 (origin) & min. & known & 0 & 0.02 & $\_$  \\
Systolic BP & 24 (origin) & min. & known & 15 & 0.03 & $\_$  \\
Diastolic BP & 17 (origin) & min. & known & 8.92 & 0.02 & $\_$  \\
Glycemia & 82 (origin) & SP & unknown & 1.18 & 0.03 & $min$  \\ \hline
\multicolumn{4}{c|}{} & $\lambda=$& 0.98 &  \\
 \cline{5-7}
\end{tabular}

\caption{Inferred model with the first dataset (898 assignment examples)
}
\label{tab:asa-phase1}
\end{table}

\paragraph*{\textbf{Second Dataset}}: in a second step, we choose to remove the 97 patients of the learning set assigned to Category 1 and whose glycemia values lie within $[0.9,1.2]$ g/l, i.e., with normal glycemia. Our goal is to foster the algorithm to retrieve a single-peaked preference for the glycemia criterion. The distribution of glycemia values in the new learning set of the remaining 801 patients is provided in Fig.
\ref{fig:hist-phase1}. 

\begin{figure}[ht!]
  \centering
  \includegraphics[scale=0.7]{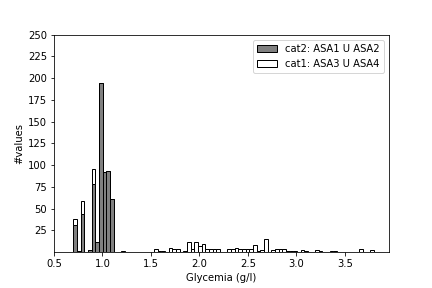}
  \label{fig:hist-phase2}
  \caption{Distribution of patients' glycemia in the second dataset 
  }
\end{figure}

We solve the inference problem with the MIP algorithm using this second learning set. Computation time is $56mn$, and the inferred model (see Table~\ref{tab:asa-phase2}) restores 99.8\% of the learning set. Once again, the restoration rate is high. However, the glycemia criterion is still detected as a cost criterion to be minimized (instead of a single-peaked criterion). The inferred model does not distinguish patients with hypoglycemia from normal glycemia ones.

\begin{table}[ht!]
\centering
\begin{tabular}{|c|ccc|>{\centering\arraybackslash}p{1.2cm}>{\centering\arraybackslash}p{1.2cm}>{\centering\arraybackslash}p{1.4cm}|} \hline 
& \multicolumn{3}{|c}{\textbf{\begin{tabular}[c]{@{}c@{}} Instance settings \end{tabular}}} & 
\multicolumn{3}{|c|}{\textbf{\begin{tabular}[c]{@{}c@{}} Model parameters learned \end{tabular}}}  \\
Attributes & $\#values$ & Direction & pref. dir.  & $b_i$ & $w_i$ & pref. dir.  \\ \hline 
Age & 103 (origin) & min. & known & 5.9 & 0 & $\_$   \\
Diabetic &  2 (origin) & min. & known & 0.99 & 0 & $\_$  \\
Hypertension &  2 (origin) & min. & known & 0 & 0.01 & $\_$  \\
Respiratory F & 2 (origin) & min. & known & 0 & 0.01 & $\_$  \\
Pacemaker & 2 (origin) & min. & known & -0.01 & 0 & $\_$  \\
Systolic BP & 23 & min. & known & 15 & 0.01 & $\_$  \\
Diastolic BP & 15 & min. & known & 8.5 & 0.01 & $\_$  \\
Glycemia & 82 (origin) & SP & unknown & 1.18 & 0.96 & $min$  \\
\hline
\multicolumn{4}{c|}{} & $\lambda=$  &0.99  &  \\
\cline{5-7}

\end{tabular}
\caption{Inferred model with the second dataset (801 assignment examples)}
\label{tab:asa-phase2}
\end{table}

\begin{figure}[ht!]
  \centering
  \includegraphics[scale=0.7]{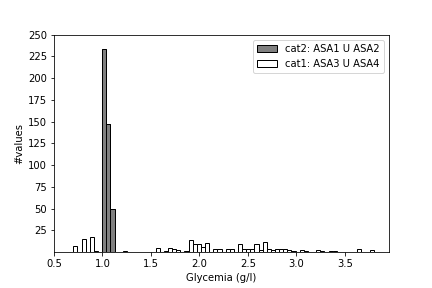}
  \caption{Patients' glycemia in the third dataset
  }
  \label{fig:hist-phase3}
\end{figure}

\paragraph{\textbf{Third Dataset}}: Finally, we remove patients in Category 2 for which the glycemia value is lower than 0.9 (hypoglycemia). This new configuration leads to a dataset of 624 patients. In this third dataset, patients suffering from hypo or hyperglycemia are assigned to Category 1 while, patients with normal glycemia are assigned to Category 2. This is illustrated by the histogram in  Figure~\ref{fig:hist-phase3}. 

With this dataset, the MIP algorithm runs in $4mn30s$ and results are presented in Table~\ref{tab:asa-phase3}. The computed model restores all the assignment examples, and glycemia is now detected as a single-peaked criterion. Furthermore, the approved values [0.93, 1.18] can be reasonably interpreted as normal glycemia.

\begin{table}[ht!]
\centering
\begin{tabular}{|c|ccc|>{\centering\arraybackslash}p{1.2cm}>{\centering\arraybackslash}p{1.2cm}>{\centering\arraybackslash}p{1.4cm}|} \hline 
& \multicolumn{3}{|c}{\textbf{\begin{tabular}[c]{@{}c@{}} Instance settings \end{tabular}}} & 
\multicolumn{3}{|c|}{\textbf{\begin{tabular}[c]{@{}c@{}} Model parameters learned \end{tabular}}}  \\
Attributes & $\#values$ & Direction & pref. dir.  & $b_i$ & $w_i$ & pref. dir.  \\ \hline 
Age          & 103 (origin)& min.  & known & 3.3  & 0 & $\_$   \\
Diabetic     &  2 (origin) & min.  & known & 0 & 0 & $\_$  \\
Hypertension &  2 (origin) & min.  & known & 0 & 0 & $\_$  \\
Respiratory F & 2 (origin) & min.  & known & 0.99 & 0 & $\_$  \\
Pacemaker    & 2 (origin)  & min.  & known & 0 & 0 & $\_$  \\
Systolic BP  & 23          & min.  & known & 12.88 & 0.01 & $\_$  \\
Diastolic BP & 15          & min.  & known & 9 & 0.01 & $\_$  \\
Glycemia     & 73 (origin) & SP    & unknown & [0.93,1.18]$\;$ & 0.99 & $SP$  \\
\hline
\multicolumn{4}{c|}{} & $\lambda=$  &1  &  \\
\cline{5-7}

\end{tabular}
\caption{Inferred model with the third dataset (624 assignment examples)}
\label{tab:asa-phase3}
\end{table}

This illustrative example shows that our model is able to infer an MR-Sort model and to retrieve single-peaked criteria; however, to do so, the learning set should be sufficiently informative.

%===============================
\section{Conclusion and future work}
\label{sec:conclusion}
%===============================

In this paper, we propose a MIP-based method to infer a MR-Sort model from a set of assignment examples when considering possible non monotone preferences. More precisely we learn an Mr-Sort model with criteria that can either be of type \textit{(i)} cost, \textit{(ii)} gain, \textit{(iii)} single-peaked or \textit{(iv)} single-valley criteria. Our inference procedure simultaneously infers from the dataset an MR-Sort model and the type of each criterion.\\[-2mm] 

Our experimental test on simulated data shows that the MIP resolution makes it possible to handle datasets involving 200 examples and 9 criteria. Experiments suggest that the correct restoration of the criteria type \textit{(i)-(iv)} requires datasets of significant size. Moreover, restoration of a criterion type is even more difficult when this criterion has a low importance.\\[-2mm]

Our work opens avenues for further research. First, it would be interesting to test our methodology on real-world case studies to further assess and investigate the performance and applicability of our proposal. Another research direction aims at pushing back a computational barrier: our MIP resolution approach faces a combinatorial explosion. The design of an efficient heuristic would be beneficial in this respect.

\bibliographystyle{ecai2012}
\bibliography{biblio}

\begin{thebibliography}{10}

\bibitem{article:Belahcene2018COR}
K.~Belahcene, C.~Labreuche, N.~Maudet, V.~Mousseau, and W.~Ouerdane, `An
  efficient {SAT} formulation for learning multiple criteria non-compensatory
  sorting rules from examples', {\em Computers \& Operations Research}, {\bf
  97},  58--71, (2018).

\bibitem{black1948}
D.~Black, `On the rationale of group decision-making', {\em Journal of
  Political Economy}, {\bf 56}(1),  23--34, (1948).

\bibitem{BLASZCZYNSKI2012284}
J.~Blaszczynski, S.~Greco, and R.~Slowinski, `Inductive discovery of laws using
  monotonic rules', {\em Engineering Applications of Artificial Intelligence},
  {\bf 25}(2),  284--294, (2012).

\bibitem{bouyssou2007a}
D.~Bouyssou and T.~Marchant, `An axiomatic approach to noncompensatory sorting
  methods in mcdm, i: The case of two categories', {\em European Journal of
  Operational Research}, {\bf 178},  217--245, (February 2007).

\bibitem{bouyssou2007b}
D.~Bouyssou and T.~Marchant, `An axiomatic approach to noncompensatory sorting
  methods in {MCDM}, {II}: {More} than two categories', {\em European Journal
  of Operational Research}, {\bf 178}(1),  246--276, (2007).

\bibitem{BUTLER1997531}
J.~Butler, J.~Jia, and J.~Dyer, `Simulation techniques for the sensitivity
  analysis of multi-criteria decision models', {\em European Journal of
  Operational Research}, {\bf 103}(3),  531--546, (1997).

\bibitem{Cano2019168}
J.-R. Cano, P.A. Gutierrez, B.~Krawczyk, M.~Wozniak, and S.~Garcia, `Monotonic
  classification: An overview on algorithms, performance measures and data
  sets', {\em Neurocomputing}, {\bf 341},  168--182, (2019).

\bibitem{ibmurl}
IBM~ILOG Cplex, {\em IBM ILOG CPLEX Optimization Studio CPLEX User’s Manual,
  Version 12, Release 8}, IBM ILOG, 20.1.0 edn., 2017.

\bibitem{Despotis1995}
D.~K. Despotis and C.~Zopounidis, {\em Building Additive Utilities in the
  Presence of Non-Monotonic Preferences},  101--114, Springer, 1995.

\bibitem{Doumpos2012}
M.~Doumpos, `Learning non-monotonic additive value functions for multicriteria
  decision making', {\em {OR} Spectrum}, {\bf 34}(1),  89--106, (2012).

\bibitem{DOUMPOS2009496}
M.~Doumpos, Y.~Marinakis, M.~Marinaki, and C.~Zopounidis, `An evolutionary
  approach to construction of outranking models for multicriteria
  classification: The case of the {ELECTRE TRI} method', {\em European Journal
  of Operational Research}, {\bf 199}(2),  496--505, (2009).

\bibitem{DoumposZopounidis2002}
M.~Doumpos and C.~Zopounidis, {\em Multicriteria Decision Aid Classification
  Methods}, Kluwer Academic Publishers, Dordrecht, 2002.

\bibitem{Eckhardt2012PreprocessingAF}
A.~Eckhardt and T.~Kliegr, `Preprocessing algorithm for handling non-monotone
  attributes in the {UTA} method', in {\em Proceedings of the ECAI-12 Workshop
  on Preference Learning: Problems and Applications in AI (PL-12)}, eds.,
  J.~F{\"{u}}rnkranz and E.~H{\"{u}}llermeier, (2012).

\bibitem{Feelders10}
A.~Feelders, `Monotone relabeling in ordinal classification', in {\em {ICDM}
  2010, The 10th {IEEE} International Conference on Data Mining}, eds., G.I.
  Webb, B.~Liu, C.~Zhang, D.~Gunopulos, and X.~Wu, pp. 803--808. {IEEE}
  Computer Society, (2010).

\bibitem{figueira2005electre}
J.~Figueira, V.~Mousseau, and B.~Roy, `Electre methods', in {\em Multiple
  criteria decision analysis: State of the art surveys},  133--153, Springer,
  (2005).

\bibitem{GHADERI2017}
M.~Ghaderi, F.~Ruiz, and N.~Agell, `A linear programming approach for learning
  non-monotonic additive value functions in multiple criteria decision aiding',
  {\em European Journal of Operational Research}, {\bf 259}(3),  1073 -- 1084,
  (2017).

\bibitem{GrecoMS01}
S.~Greco, B.~Matarazzo, and R.~Slowinski, `Rough sets theory for multicriteria
  decision analysis', {\em European Journal of Operational Research}, {\bf
  129}(1),  1--47, (2001).

\bibitem{GUO2019}
M.~Guo, X.~Liao, and J.~Liu, `A progressive sorting approach for multiple
  criteria decision aiding in the presence of non-monotonic preferences', {\em
  Expert Systems with Applications}, {\bf 123},  1 -- 17, (2019).

\bibitem{GutierrezG16}
P.A. Guti{\'{e}}rrez and S.~Garc{\'{\i}}a, `Current prospects on ordinal and
  monotonic classification', {\em Prog. Artif. Intell.}, {\bf 5}(3),  171--179,
  (2016).

\bibitem{KADZINSKI202060}
M.~Kadzinski, K.~Martyn, M.~Cinelli, R.~Slowinski, S.~Corrente, and S.~Greco,
  `Preference disaggregation for multiple criteria sorting with partial
  monotonicity constraints: Application to exposure management of
  nanomaterials', {\em International Journal of Approximate Reasoning}, {\bf
  117},  60--80, (2020).

\bibitem{Kadzinski2021a}
M.~Kadzinski, K.~Martyn, M.~Cinelli, R.~Slowinski, S.~Corrente, and S.~Greco,
  `Preference disaggregation method for value-based multi-decision sorting
  problems with a real-world application in nanotechnology', {\em
  Knowledge-Based Systems}, {\bf 218}, (2021).

\bibitem{KADZINSKI2020}
M.~Kadzinski and M.~Martyn, `Enriched preference modeling and robustness
  analysis for the {ELECTRE Tri-B} method', {\em Annals of Operations
  Research},  1--35, (2020).

\bibitem{Kliegr2009UTANME}
T.~Kliegr, `{UTA-NM} : Explaining stated preferences with additive
  non-monotonic utility functions', in {\em Proceedings of ECML PKDD Workshop
  on Preference Learning}, (2009).

\bibitem{Lazouni_2013}
M.A. Lazouni, M.A. Chikh, and S.~Mahmoudi, `A new computer aided diagnosis
  system for pre-anesthesia consultation', {\em Journal of Medical Imaging and
  Health Informatics}, {\bf 3}(4),  471--479, (2013).

\bibitem{leroy2011learning}
A.~Leroy, V.~Mousseau, and M.~Pirlot, `Learning the parameters of a multiple
  criteria sorting method', in {\em International Conference on Algorithmic
  Decision Theory}, pp. 219--233. Springer, (2011).

\bibitem{LIU20191071}
J.~Liu, X.~Liao, M.~Kadzinski, and R.~Slowinski, `Preference disaggregation
  within the regularization framework for sorting problems with multiple
  potentially non-monotonic criteria', {\em European Journal of Operational
  Research}, {\bf 276}(3),  1071--1089, (2019).

\bibitem{MeyerOlteanu_2017}
P.~Meyer and A.~Olteanu, `Integrating large positive and negative performance
  differences into multicriteria majority-rule sorting models', {\em Computers
  and Operations Research}, {\bf 81},  216–230, (2017).

\bibitem{MeyerOlteanu_2019}
P.~Meyer and A.~Olteanu, `Handling imprecise and missing evaluations in
  multi-criteria majority-rule sorting', {\em Computers and Operations
  Research}, {\bf 110},  135–147, (2019).

\bibitem{minoungouDA2PL2020}
P.~Minoungou, V.~Mousseau, W.~Ouerdane, and P.~Scotton, `{Learning an MR-sort
  model from data with latent criteria preference direction}', in {\em
  {DA2PL'2020, from multiple criteria Decision Aid to Preference Learning}},
  Trento, Italy, (2020).

\bibitem{mousseau1998}
V.~Mousseau and R.~Slowinski, `Inferring an {ELECTRE} {TRI} model from
  assignment examples', {\em Journal of global optimization}, {\bf 12}(2),
  157--174, (1998).
\newblock Springer.

\bibitem{Nefla_2019}
O.~Nefla, M.~Öztürk, P.~Viappiani, and I.~Brigui-Chtioui, `Interactive
  elicitation of a majority rule sorting model with maximum margin
  optimization', in {\em ADT 2019, 6th International Conference on Algorithmic
  Decision Theory}, (2019).

\bibitem{Pei2018104}
S.~Pei and Q.~Hu, `Partially monotonic decision trees', {\em Information
  Sciences}, {\bf 424},  104--117, (2018).

\bibitem{Roy1991}
B.~Roy, `The outranking approach and the foundations of {E}lectre methods',
  {\em Theory and Decision}, {\bf 31}(1),  49--73, (1991).

\bibitem{sobrie2016}
O.~Sobrie, {\em Learning preferences with multiple-criteria models}, Ph.D.\
  dissertation, Université de Mons (Faculté Polytechnique) and Université
  Paris-Saclay (CentraleSupélec), 2016.

\bibitem{sobrie2016ASA}
O.~Sobrie, M.A. Lazouni, S.~Mahmoudi, V.~Mousseau, and M.~Pirlot, `A new
  decision support model for preanesthetic evaluation', {\em Computer Methods
  and Programs in Biomedicine}, {\bf 133},  183--193, (2016).

\bibitem{SobrieMousseauPirlot2019}
O.~Sobrie, V.~Mousseau, and M.~Pirlot, `Learning monotone preferences using a
  majority rule sorting model', {\em Int. Trans. Oper. Res.}, {\bf 26}(5),
  1786--1809, (2019).

\bibitem{Wang2015172}
H.~Wang, M.~Zhou, and K.~She, `Induction of ordinal classification rules from
  decision tables with unknown monotonicity', {\em European Journal of
  Operational Research}, {\bf 242}(1),  172--181, (2015).

\bibitem{zheng2014COR}
J.~Zheng, S.~Metchebon~Takougang, V.~Mousseau, and M.~Pirlot, `Learning
  criteria weights of an optimistic electre tri sorting rule', {\em Computers
  and Operations Research}, {\bf 49},  28 -- 40, (2014).

\bibitem{zopoudoupos-review}
C.~Zopounidis and M.~Doumpos, `{Multicriteria classification and sorting
  methods: A literature review}', {\em European Journal of Operational
  Research}, {\bf 138}(2),  229--246, (2002).

\end{thebibliography}

\end{document}